\documentclass[letterpaper, 10 pt, conference]{ieeeconf}

\IEEEoverridecommandlockouts           
 
\usepackage[pdftex]{graphicx}
\usepackage{amssymb}
\usepackage{amsmath}
\usepackage{bm}
\usepackage{cite}
\usepackage{epsfig}
\usepackage{caption}
\usepackage{color}
\usepackage{psfrag}
\usepackage{cases}
\usepackage{acronym}
\usepackage{setspace}
\usepackage{times}
\usepackage{latexsym}
\usepackage{dblfloatfix}
\usepackage{subcaption}
\usepackage{metalogo}
\usepackage{tabularx}
\usepackage{array}
\usepackage{multirow}
\usepackage{booktabs}
\usepackage{url}

\title{Semantic Graph Based Place Recognition for 3D Point Clouds}

\author{Xin Kong$^{1}$, Xuemeng Yang$^{1}$, Guangyao Zhai$^{1}$, Xiangrui Zhao$^{1}$,\\Xianfang Zeng$^{1}$, Mengmeng Wang$^{1}$, Yong Liu$^{1}$$^{*}$, Wanlong Li$^{2}$ and Feng Wen$^{2}$\\
		\thanks{$^{1}$The authors are with the Institute of Cyber-Systems and Control, Zhejiang University, Hangzhou, 310027, China. (Yong Liu$^{*}$ is the corresponding author, email: yongliu@iipc.zju.edu.cn)}%
		\thanks{$^{2}$The authors are with Huawei Noah's Ark Lab, Beijing, China.}
		\thanks {This work is supported by the National Natural Science Foundation of China under Grant 61836015. We thank Huan Yin for fruitful discussion.}
}

\begin{document}

\maketitle
\thispagestyle{empty}
\pagestyle{empty}

\begin{abstract}
Due to the difficulty in generating the effective descriptors which are robust to occlusion and viewpoint changes, place recognition for 3D point cloud remains an open issue. Unlike most of the existing methods that focus on extracting local, global, and statistical features of raw point clouds, our method aims at the semantic level that can be superior in terms of robustness to environmental changes. Inspired by the perspective of humans, who recognize scenes through identifying semantic objects and capturing their relations, this paper presents a novel semantic graph based approach for place recognition. First, we propose a novel semantic graph representation for the point cloud scenes by reserving the semantic and topological information of the raw point cloud. Thus, place recognition is modeled as a graph matching problem. Then we design a fast and effective graph similarity network to compute the similarity. Exhaustive evaluations on the KITTI dataset show that our approach is robust to the occlusion as well as viewpoint changes and outperforms the state-of-the-art methods with a large margin. Our code is available at: \url{https://github.com/kxhit/SG_PR}.
   
\end{abstract}

\section{Introduction}

In the past few decades, Simultaneous Localization and Mapping (SLAM) has developed rapidly, which plays a critical role in robotic applications and autonomous driving. Loop closure detection is an important issue in SLAM, which refers to the ability of robots or moving vehicles to recognize whether a place has been reached before. It is the most effective way to eliminate the cumulative odometry drift error, helping build a more precise global map and achieve more accurate localization. Current strategies for place recognition are primarily based on descriptors generation and feature distance measurement. 

\begin{figure}[h]
	\centerline{\includegraphics[height=5cm]{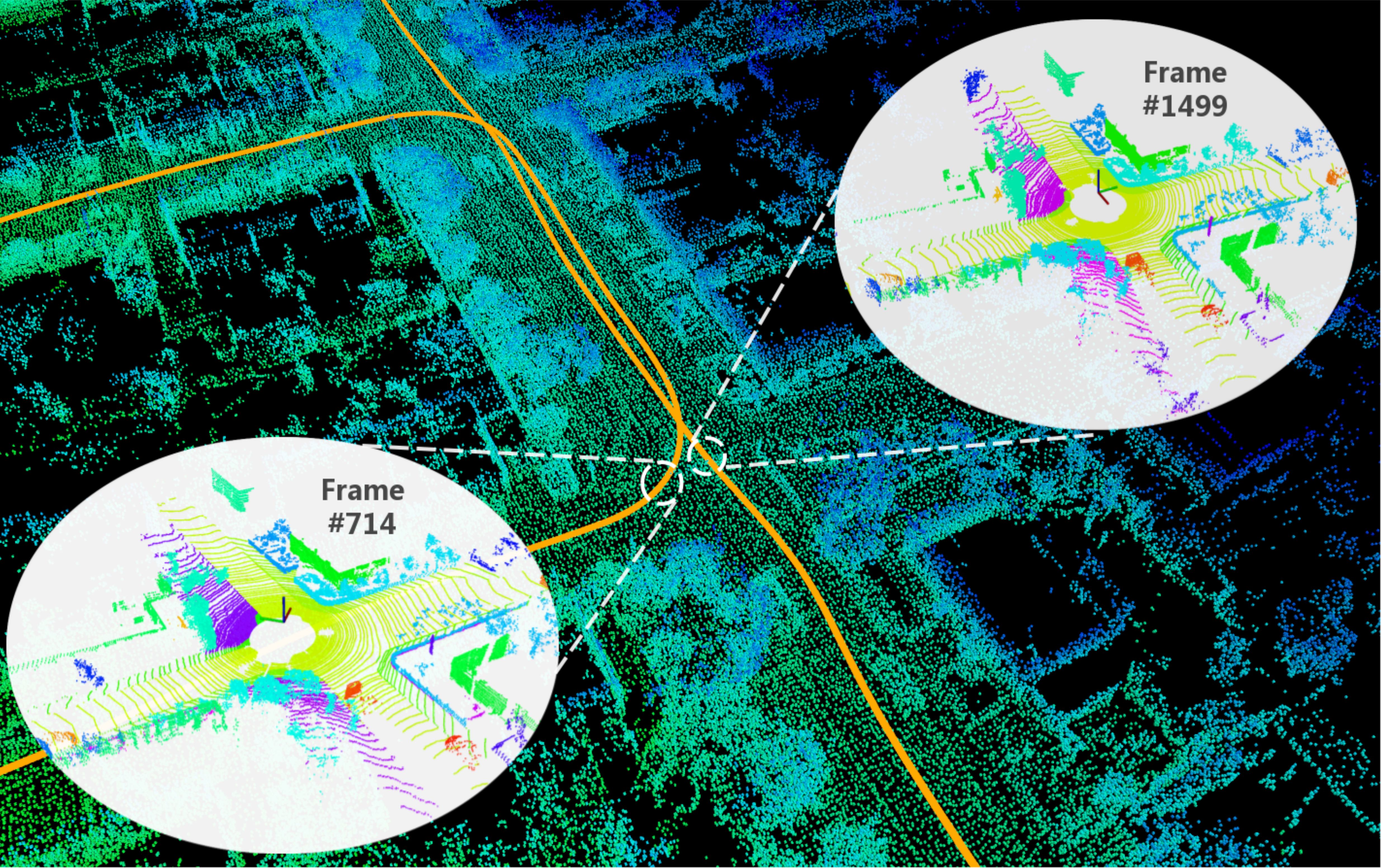}}
	\caption{An illustration of place recognition. This is a reverse loop closure in sequence 08 of KITTI detected by our proposed approach. Note that the heading direction of frame 714 and 1499 are almost exactly the opposite, which brings a challenge to existing methods. (Best viewed with zoom-in.)} 
	\label{fig:demo}
\end{figure}

Research on vision-based place recognition has been investigated for a long time and many successful approaches have been proposed~\cite{cummins2008fab,milford2012seqslam,sattler2016large}. Most of the image-based methods extract feature descriptors and then encode them with methods such as bag-of-words (BoW)~\cite{cummins2008fab, lazebnik2006beyond}, VLAD~\cite{jegou2010aggregating} and Fisher Vector (FV)~\cite{dixit2015scene, dixit2016object}. The relevant scenes are retrieved by comparing the global descriptors and measuring the similarity among them. However, due to the interference of external conditions such as weather, seasons, illumination, and viewpoint changes, image-based methods are probably failed to retrieve the correct match~\cite{lowry2015visual}.

\begin{figure*}[ht]
	\centerline{\includegraphics[height=7.5cm]{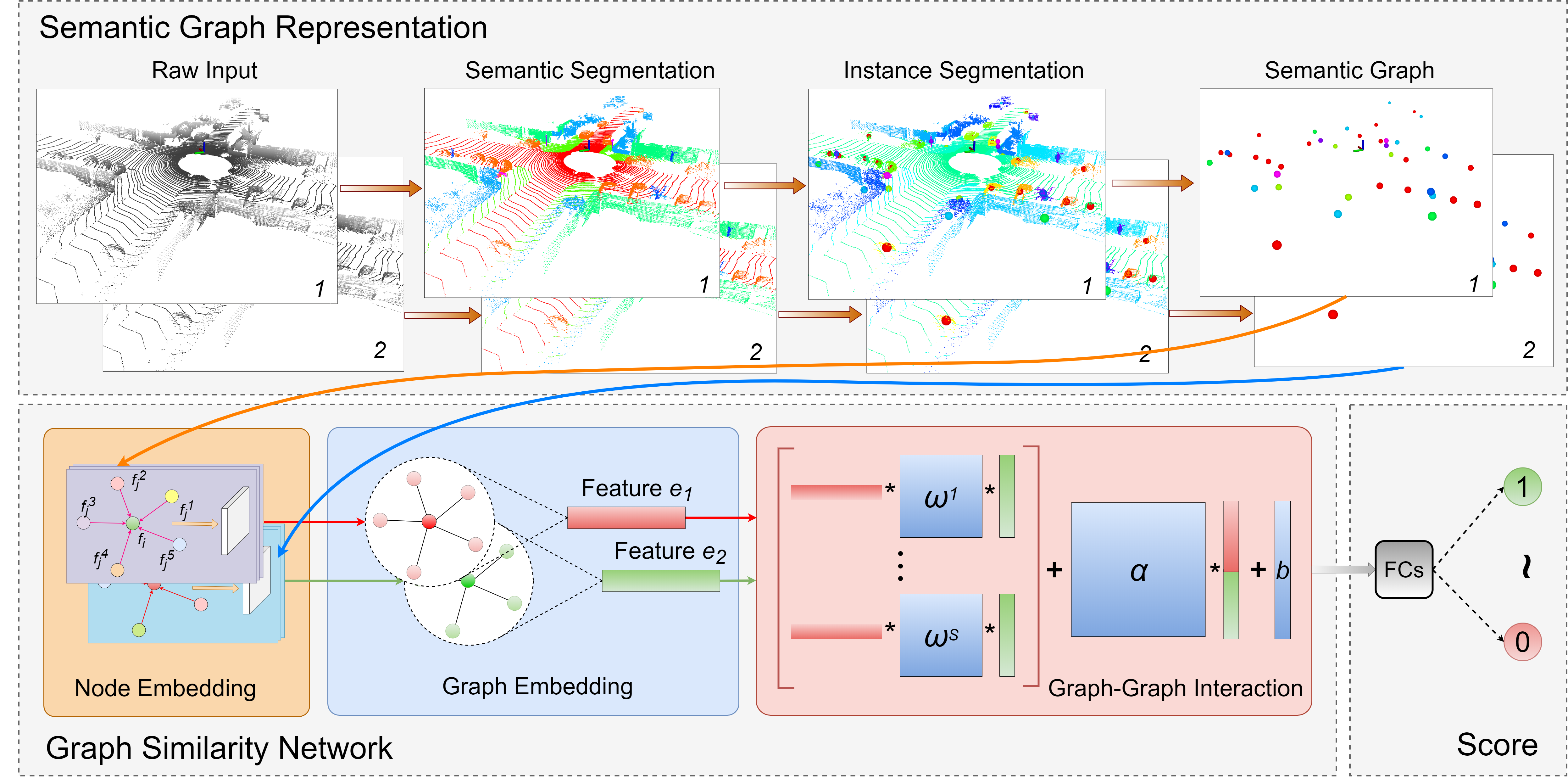}}
	\caption{The architecture of our approach. The whole framework consists of semantic graph representation and graph similarity network. We first segment the point clouds semantically to obtain instances on raw point clouds. In addition, we aggregate the semantic and topological information to acquire nodes and form the semantic graph. By feeding pairs of graphs into graph similarity network composed of node embedding, graph embedding, and graph-graph interaction, we acquire a similarity score in the range $[0, 1]$. FCs denotes a set of fully connected layers.} 
	\label{fig:pipeline}
\end{figure*}

LiDAR-based methods are recently gained widespread attention, as they are more robust to seasons and illumination variations. Most LiDAR-based algorithms~\cite{he2016m2dp, angelina2018pointnetvlad, liu2019lpd, liu2019seqlpd} operate directly on raw point cloud data and generate local or global descriptors by neural networks or handcrafted design. Such methods obtain 
low-level features like local structures and distributing characteristics of points, 
which are sensitive to occlusion and rotation. A few segment based approaches~\cite{dube2017segmatch,segmap2018,dube2019segmap} recognize places by matching segments that belong to partial or full objects, which can better represent dynamic situations. These methods are 
more related to the way humans perceive their surroundings. However, it's hard to obtain the accurate and stable feature of segments and they ignore relations among segments, which is 
crucial to the scene expression. 

Humans perceive the environment by distinguishing scenes through semantic objects and their topological relations. Inspired by this, we present a new approach that converts the raw point cloud data to a novel graph representation by aggregating the semantic information. Such graph representation retains critical information and considers the topological relations, making the expression of the point cloud data more efficient and comprehensible. Moreover, we apply a learning-based graph similarity computation strategy to solve the retrieve task instead of simply calculating the Euclidean distances of feature vectors. To the best of our knowledge, we are the first to use semantic graph representation and graph matching for place recognition in 3D point clouds. A demonstration of our results is displayed in Fig.~\ref{fig:demo}. Our contributions can be summarized as follows:

(1) Towards humanoid perception, we present a novel semantic graph representation for 3D point cloud scenes, which captures semantic information and models topological relations between semantic objects.
		
(2) We propose an effective and efficient network to estimate the graph matching similarity among point cloud scenes which can be used in loop closure detection.
	
(3) Experiments on the KITTI odometry dataset~\cite{geiger2013vision} show that our approach achieves state-of-the-art performance, especially for reverse loop closure detection and the robustness to occlusion as well as viewpoint changes.

\section{Related Work}

The loop closure detection methods based on 3D point cloud can be divided into the following categories: local descriptor based methods, global descriptor based methods, and segments based methods. 

\smallskip
\textbf{Local descriptor based methods:} Spin image\cite{johnson1999using} first generates a cylindrical coordinate system around each key point, separates the nearby points into bins, and then encodes a pattern of surrounding bins into a histogram. Bosse and Zlot~\cite{bosse2013place} query a constant number of nearest neighbor votes for each keypoint from the database of local 3D Gestalt descriptors and places with a sufficient number of votes are determined as possible location matches. Whereas such local keypoint features often lack descriptive power to distinguish similar local structures and are not always reliable for matching.

\smallskip
\textbf{Global descriptor based methods:}	ESF~\cite{wohlkinger2011ensemble} presents a global shape descriptor using a concatenation of histograms describing distance, angle, and area distributions on the surface of the partial point cloud. Without extracting the normal vectors of each point, the lack of spatial information in these descriptors makes it hard to capture intricate details in different clouds. M2DP~\cite{he2016m2dp} projects a 3D point cloud onto multiple 2D planes and generates density signatures. The left and right singular vectors of these signatures are then used as descriptors for the 3D scene. However, it relies on the distribution of all points and the performance is not satisfied when there is a partial loss of points. LiDAR Iris~\cite{wang2019lidar} extracts corresponding binary signature images from point clouds and measure similarities by calculating hamming-distance. PointNetVLAD~\cite{angelina2018pointnetvlad} combines PointNet~\cite{qi2017pointnet} and NetVLAD~\cite{arandjelovic2016netvlad}, which is the first point cloud network to directly handle the point cloud scenes in 3D space. But the local feature extraction and the spatial distribution of local features are not fully considered. SeqLPD~\cite{liu2019seqlpd} and LPD-Net~\cite{liu2019lpd} extract features in both feature space and Cartesian space, fuse the neighborhood features of each point, and use NetVLAD to generate the global descriptors. The above methods process a large number of raw points and achieve unsatisfactory performance when the point cloud scenes rotate. Scan Context~\cite{Kim2018ScanCE} reserves maximum heights and maps 3D point clouds to 2D planes by histogram statistics. To achieve rotation invariance, it calculates all possible column-shifted scan contexts line-by-line to find the minimum distance, requiring longer search time than others.

\smallskip
\textbf{Segments based methods:} SegMatch~\cite{dube2017segmatch} and SegMap~\cite{segmap2018} present a high-level perception which segments point clouds into a set of distinct and discriminative elements at object-level. They use a 3D CNN to encode segment features and identify candidate correspondences by using k nearest neighbors (kNN) in feature space. This approach is a successful attempt towards humanoid perception. Nonetheless, it needs a dense local map, and relations among objects are not taken into consideration.

To address the above problems, we create a novel graph representation at a semantic level, making it more concise and effective. Then we apply a graph similarity network instead of Euclidean distance to measure  similarities of scenes for better estimation.

\section{Methodology}

In this section, we present our semantic graph based place recognition approach, which consists of semantic graph representation and learning-based graph similarity computation, as shown in Fig.~\ref{fig:pipeline}. Our key insight is to recognize the scenes from the human perspective, describe it at the semantic level, and focus on encoding relations among semantic objects. People usually recognize scenes by identifying some semantic objects and observing their relative positions. For this reason, we  utilize semantic segmentation on raw point clouds to obtain instances and further collect semantic and topological information together to acquire nodes forming the semantic graph. After this, the raw point cloud scene is transformed into a topological semantic graph and the place recognition task is now converted to a graph matching problem. Moreover, a learning-based graph similarity computation is applied to obtain the similarity scores between pairs of scenes.

To the best of our knowledge, this is the first attempt to use topological semantic graph representation and graph matching in loop closure detection.

\subsection{Semantic Graph Representation}

\begin{figure}[t]
	\centerline{\includegraphics[height=3.7cm]{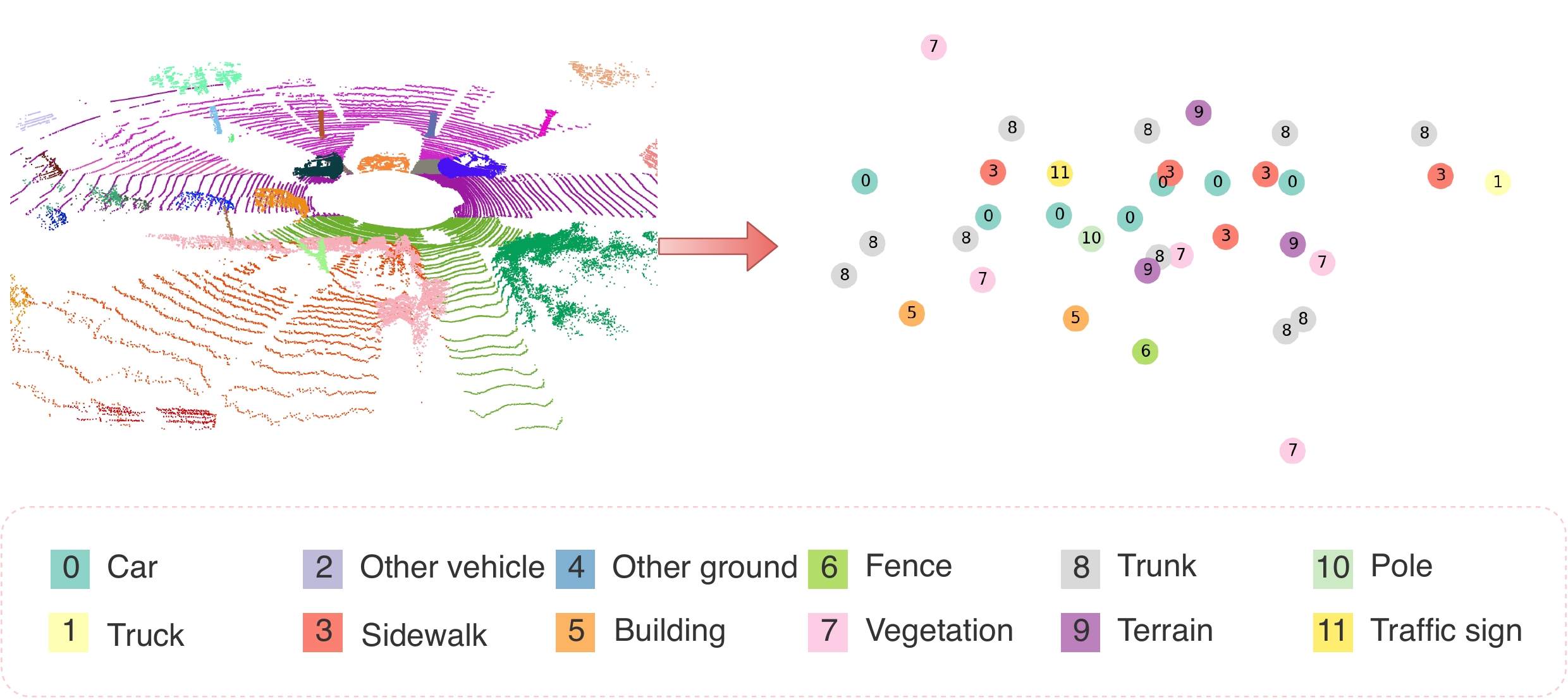}}
	\caption{An illustration of semantic graph representation for point clouds. Each node denotes an instance in the scene, reserving its semantic and topological information.} \label{fig:node_representation}
\end{figure}

\smallskip
\textbf{Semantic Segmentation for Point Cloud:} Some semantic segmentation methods for point cloud have been proposed recently~\cite{Wu2017SqueezeSegCN, wu2019squeezesegv2, Kong2019PASS3DPA, milioto2019rangenet++}. RangeNet++~\cite{milioto2019rangenet++} is trained on SemanticKITTI~\cite{behley2019semantickitti} dataset, which annotates the semantic categories of each 3D point on the KITTI~\cite{geiger2013vision} odometry dataset, including a total of 19 classes. In the experimental part, we use predictions of RangeNet++ (which can be flexibly replaced by other methods) and annotations of SemanticKITTI as semantic information respectively. We merge dynamic classes to their corresponding static ones and ignore classes like person, because they are either irrelevant or few in number. The merged 12 categories are shown in Fig.~\ref{fig:node_representation}. Then, we set different clustering radii according to semantic categories and obtain semantic instances by Euclidean clustering. Specifically, for one single frame of point cloud $P=\left\{p_{1}, \dots, p_{M} | p_{i} \in \mathbb{R}^{3}\right\}$, we acquire the semantic label of each point $p_i$, and cluster points with the same semantic label into a set of clusters $I=\left\{I_{1}, \dots, I_{N}\right\}$, $I_i=\left\{p_{1}, \dots, p_{} | p_{i} \in \mathbb{R}^{3}\right\} \subset I$ which indicates different instances, and their corresponding semantic labels are $L=\left\{l_{1}, \dots, l_{N}| l_{i} \in \mathbb{R}^{1}\right\}$.

\smallskip
\textbf{Semantic Graph Construction:} A 64-ring LiDAR usually captures more than 100k points per frame, which is huge and redundant. To reduce the data, most of the existing methods downsample the points randomly~\cite{he2016m2dp, angelina2018pointnetvlad} or project them onto a 2D plane~\cite{yin2018locnet, Kim2018ScanCE}. Distinctively, we construct topological semantic graph representation to reserve the key information by retaining the semantic information and topological relations of the semantic instances, which is more concise and meaningful.

As shown in Fig.~\ref{fig:node_representation}, for each instance $I_i$, we represent it with a node $V_i$, preserving its semantic category $l_i$ and the centroid $\left( \bar{x}_i, \bar{y}_i, \bar{z}_i \right) $ of the set of points $\left(x_{j}, y_{j}, z_{j}\right) \in I_i$, composing the node feature $f_i\in\mathbb{R}^{4}$. In the node embedding part, the semantic feature is one-hot encoded (e.g. all the nodes with car type share the same one-hot encoding vector). Thus, these nodes together form a graph that can represent a point cloud scene. The LiDAR-based loop closure detection is determined by comparing the similarity of two scenes, which has now been turned into a similarity measurement problem for two topological semantic graphs.

\begin{figure*}[htb]
	\centerline{\includegraphics[height=5cm]{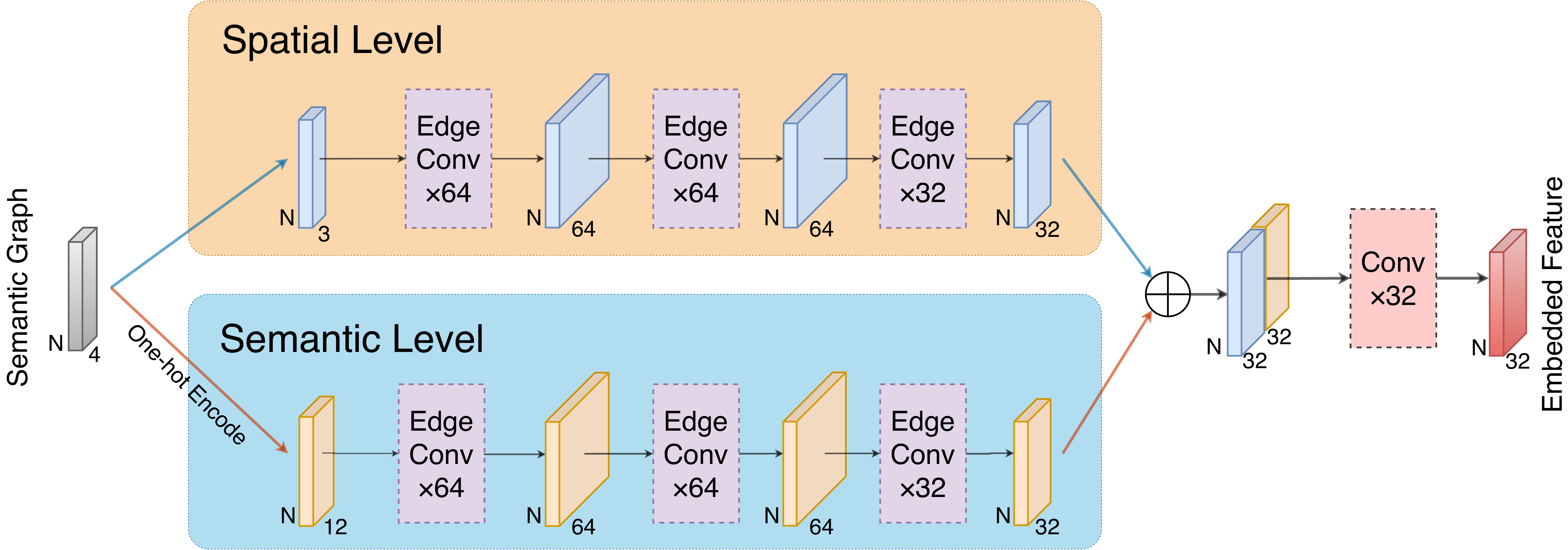}}
	\caption{Detailed architecture of node embedding. We obtain node features on spacial and semantic level respectively, and concatenate them together to get the node embedding.} 
	\label{fig:node_embedding}
\end{figure*}

\subsection{Graph Similarity Network}

Graphs have a wide range of applications and there are different similarity metrics, such as Graph Edit Distance (GED)~\cite{bunke1983distance}, Maximum Common Subgraph (MCS)~\cite{bunke1998graph}. However, computing these metrics between two graphs is NP-complete~\cite{bunke1998graph, zeng2009comparing} and it is hard to compute the exact distance within reasonable time~\cite{blumenthal2018exact}. Beyond this, some challenges need to be tackled as well when implementing graph matching in loop closure detection. The algorithm needs to be representation invariant as the computed similarity score should be permutation invariant to the order of nodes. Besides, it should be rotation invariant because reverse loop closure is common in real-world applications. The computational efficiency and generalization ability are crucial as well. Based on the above considerations, we propose the graph similarity network to perform graph matching for place recognition inspired by SimGNN~\cite{bai2019simgnn}. Our network structure is shown in Fig.~\ref{fig:pipeline}.

\smallskip
\textbf{Node Embedding:} Graph Convolutional Network (GCN)~\cite{kipf2017semi} is the most popular method for aggregating node features. However, the adjacent matrix should be defined and fixed in advance. Given that our nodes can be seen as superpoints and Dynamic Graph CNN (DGCNN)~\cite{wang2019dynamic} is effective in point cloud feature learning, we adopt EdgeConv introduced in DGCNN to capture local geometric information while guaranteeing permutation invariance. Our node embedding module applies a dynamically updated graph instead of a fixed graph to group nodes semantically. The detailed architecture is shown in Fig.~\ref{fig:node_embedding}.

In the EdgeConv layer, we find a set of k nearest neighbors (kNN) for each node $V_i$ in feature and Euclidean space, and aggregate features within each set. Each node feature $f_i$ is initialized with centroid information and one-hot encoding based on the semantic label $l_i$. Each edge represents the feature relation of $f_i$ and its k-nearest neighbors $f_j^m$, $m=1, 2, ..., k$ in feature space and edge function is defined as
\begin{equation}\label{edge_func}
h_{\Theta}\left({f}_{i}, {f}_{j}\right)=\bar{h}_{\Theta}\left({f}_{i}, {f}_{i}-{f}_{j}^{m}\right),
\end{equation}
where $\Theta$ is a series of learnable parameters. This operation combines the global information captured by $f_i$ and the local relations captured by $f_i-f_j^m$. As independent convolution is more efficient for multimodal features~\cite{husain2016combining, bao2019spatiotemporal, geng2019spatiotemporal}, we perform feature aggregation separately on the spatial and semantic level, and concatenate the output features together as the final embedding $u_i$ of each node. 

\smallskip
\textbf{Graph Embedding:} A weighted or unweighted average of node embedding is usually used for generating a graph embedding. Inspired by SimGNN~\cite{bai2019simgnn}, we would like to estimate a learnable weight matrix associated with each node by an attention module. We let the neural network learn which node should receive higher attention and be more representative of the overall graph. 

For each node $V_i$, $u_i\in\mathbb{R}^D$ indicates the node embedding of $V_i$, where $D$ is the dimension of $u_i$. We initialize the global graph context $c\in\mathbb{R}^D$ as a simple average of each node embedding and followed by an activation function, where we use a $tanh()$. So the global graph context $c$ is
\begin{equation}\label{eq:global_context}
c=tanh \left(\left(\frac{1}{N} \sum_{i=1}^{N} u_{i}\right) W\right),
\end{equation}
where $W\in\mathbb{R}^{D\times{D}}$ is a learnable weight matrix, $N$ is the number of nodes in a graph. The context $c$ provides the global structure and feature information of the graph which is updated by learning $W$. We suppose that the node similar to the global context should get higher attention weights. To make the attention $a_i$ aware of $c$, we calculate the inner product of $c$ and each node embedding $u_i$. To ensure the attention weights are in the range $[0, 1]$, we apply a sigmoid function on $a_i$. Finally, we acquire the graph embedding $e$ by the weighted sum of node embeddings as
\begin{equation}\label{eq:attention}
\begin{split}
e&=\sum_{i=1}^{N} sigmoid\left({a}_{i}\right) {u}_{i}\\
&=\sum_{i=1}^{N} sigmoid\left({u}_{i} tanh \left(\left(\frac{1}{N} \sum_{m=1}^{N} u_{m}\right) {W}\right)^{T}\right) {u}_{i}.
\end{split}
\end{equation}

\smallskip
\textbf{Graph-Graph Interaction:} The relation of a pair of graph-level embeddings $e_1$, $e_2$ can be accurately estimated by a Neural Tensor Network (NTN) presented in~\cite{socher2013reasoning}. The NTN adopts a bilinear tensor layer instead of standard linear neural network layer that directly relates the two vectors across multiple dimensions, which is a more reasonable way than calculating the inner product of $e_1$ and $e_2$. As shown in Fig.~\ref{fig:pipeline}, the model computes a feature vector to measure the relation between graph-level embeddings using a function as
\begin{equation}\label{NTN}
g\left(e_{1}, e_{2}\right)=ReLU\left(e_{1}^{T} \omega^{[1: S]} e_{2}+\alpha\left[\begin{array}{l}{e_{1}} \\ {e_{2}}\end{array}\right]+b\right),
\end{equation}
where $g\in\mathbb{R}^S$ is the output of the network. $e_1$, $e_2\in\mathbb{R}^D$ are the feature embedding of two graphs. $\omega^{[1:S]}\in\mathbb{R}^{D\times{D}\times{S}}$ represents a weight tensor, $\alpha\in\mathbb{R}^{S\times2D}$ represents a weight vector and $b\in\mathbb{R}^S$ represents a bias. The hyper-parameter $S$ denotes the number of slices and is set to 16.

\smallskip
\textbf{Graph Similarity:} We apply a set of fully connected layers to gradually reduce the dimension of the similarity vector and finally get one score per pair in the range $[0, 1]$. The ground truth is either 0 or 1 as we simplify the problem to a binary classification task. We train the model with a binary cross-entropy loss function.

\begin{figure*}[t]
	\centering
	
	\begin{subfigure}{0.3\linewidth}
		\centering
		\includegraphics[width=2.25in]{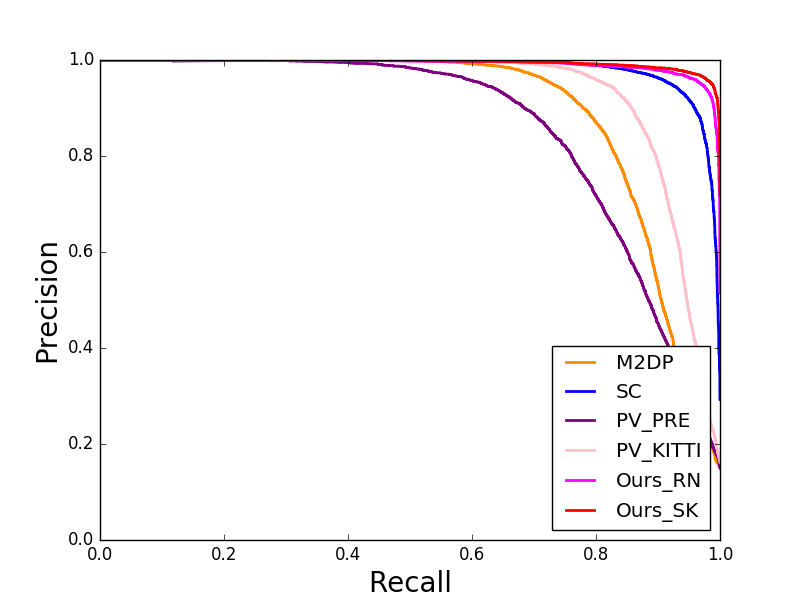}
		\caption{KITTI00}
		\label{fig:KITTI00_3m}
	\end{subfigure}
	\begin{subfigure}{0.3\linewidth}
		\centering
		\includegraphics[width=2.25in]{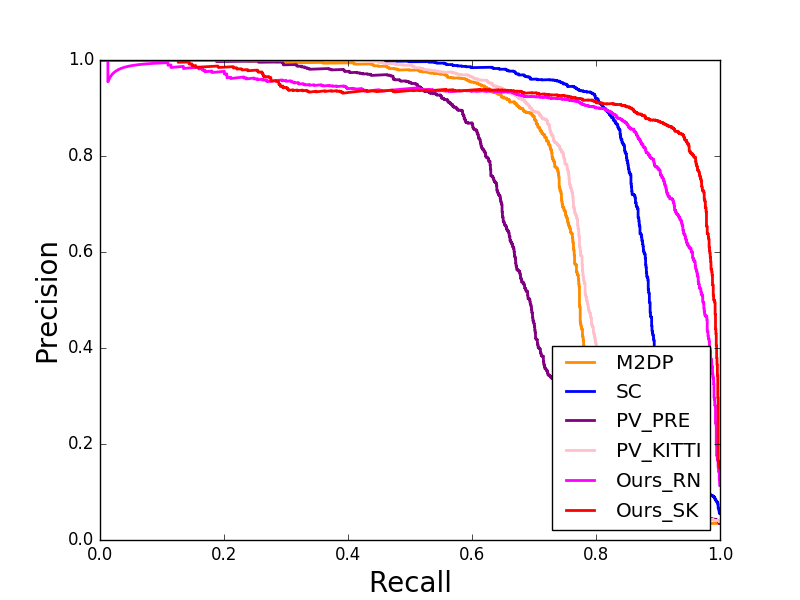}
		\caption{KITTI02}
	\end{subfigure}
	\begin{subfigure}{0.3\linewidth}
		\centering
		\includegraphics[width=2.25in]{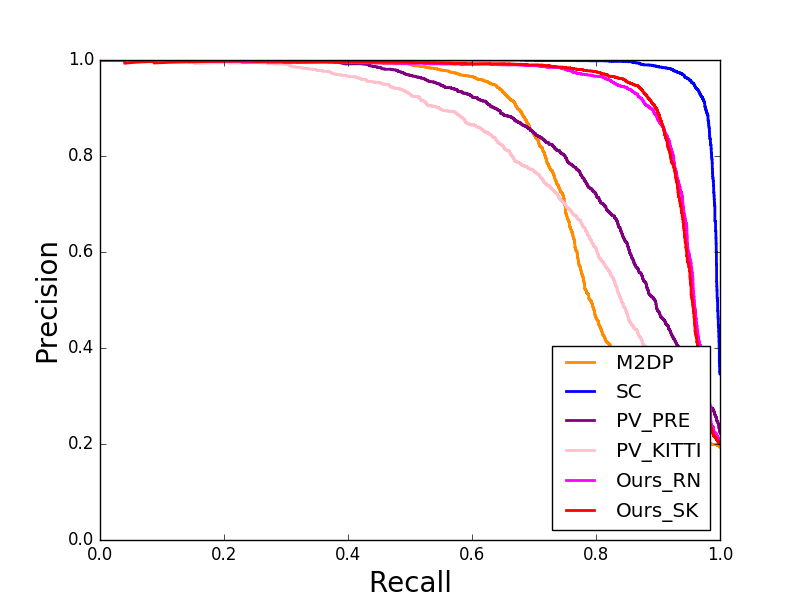}
		\caption{KITTI05}
	\end{subfigure}
	
	\begin{subfigure}{0.3\linewidth}
		\centering
		\includegraphics[width=2.25in]{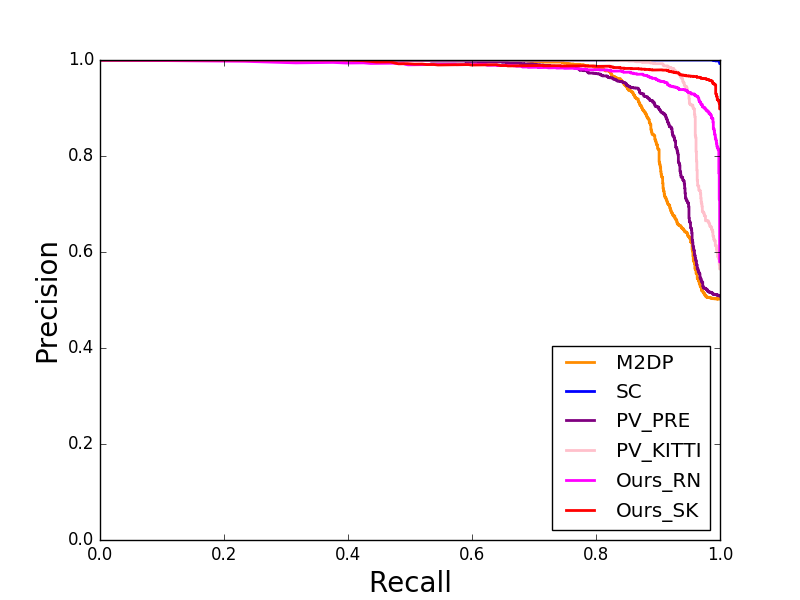}
		\caption{KITTI06}
	\end{subfigure}
	\begin{subfigure}{0.3\linewidth}
		\centering
		\includegraphics[width=2.25in]{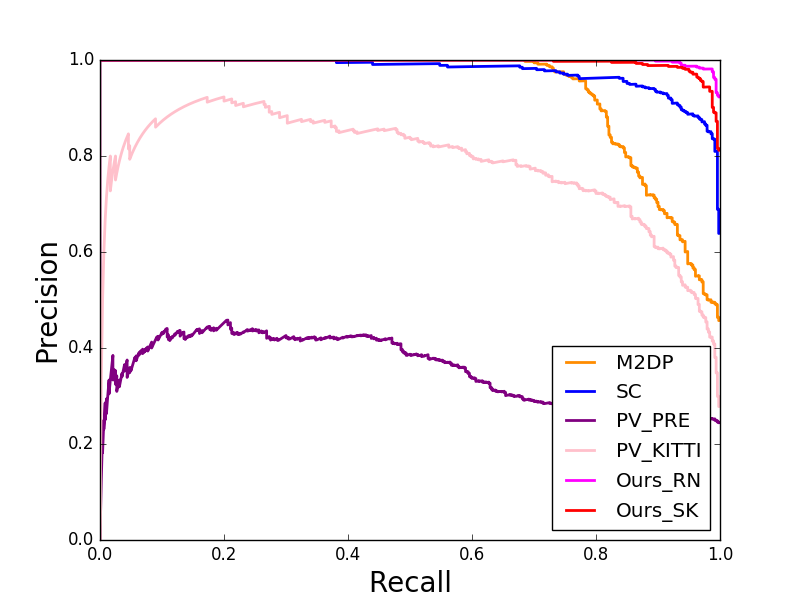}
		\caption{KITTI07}
	\end{subfigure}
	\begin{subfigure}{0.3\linewidth}
		\centering
		\includegraphics[width=2.25in]{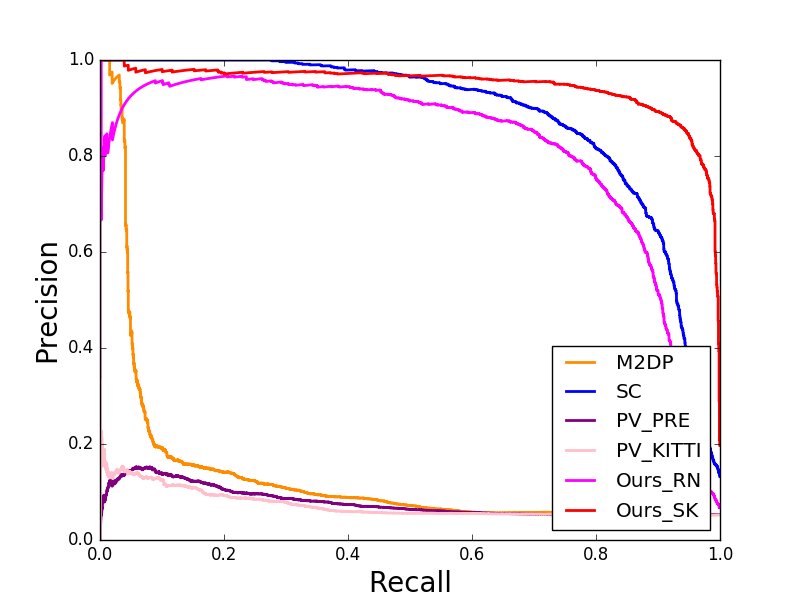}
		\caption{KITTI08}
	\end{subfigure}
	
	\centering
	\caption{Precision-Recall curves on KITTI dataset.}
	\label{Fig:PRcurve}
\end{figure*}

\section{Experiments} 

\subsection{Dataset and Implementation Details}

We evaluate the proposed method over KITTI odometry dataset~\cite{geiger2013vision}, which contains 11 sequences (from 00 to 10) obtained by a 64-ring LiDAR (Velodyne HDL-64E) with ground-truth poses. The ground-truth poses are used to determine if there is a loop closure. In our experiment, two point cloud scenes consist of a positive pair if the Euclidean distance between them is less than 3 m, while the negative one is over 20 m. Note that in the evaluation, positive pairs with a timestamp greater than 30 s are considered to be the true loop closures. In this setting, easy positive pairs (adjacent scenes) will not be evaluated thus it can reflect the real performances of the algorithms. These sequences (00, 02, 05, 06, 07, and 08) with loop closures are evaluated. The sequence 08 has reverse loops and others only have loop closures in the same direction.

The SemanticKITTI~\cite{behley2019semantickitti} dataset has 28 categories and we merge them to 12 categories as shown in Fig~\ref{fig:node_representation}. The number of nodes varies in different scenarios, ranging from 10-70 on the KITTI odometry dataset. In the node embedding part, we set $k=10$ in kNN and pad fake nodes with zero embeddings to obtain a fixed number of nodes (in our experiment, we set it to 100), thus we can train the model with a batch operation. We use 1-fold cross-validation and each sequence is considered as a fold, that is, consider one sequence as a test set and the others as training sets. All the experiments are implemented based on PyTorch~\cite{paszke2019pytorch} and Adam optimizer\cite{kingma2014adam} with a learning rate of 0.001 is used for training. There are a large number of negative pairs, thus we reserve all positive pairs and randomly sample some proportion of negative ones.

\begin{table} [t]
	\small
	\centering
	\setlength{\tabcolsep}{1mm}
	\begin{tabular}{cccccccc}
		\toprule
		Methods & 00 & 02 & 05 & 06 & 07 & 08 & Mean \cr
		\midrule
		M2DP~\cite{he2016m2dp}&0.836&0.781&0.772&0.896&0.861&0.169&0.719\cr
		SC~\cite{Kim2018ScanCE}&0.937&0.858&\textbf{0.955}&\textbf{0.998}&0.922&0.811&0.914\cr
		PV-PRE~\cite{angelina2018pointnetvlad}&0.785&0.710&0.775&0.903&0.448&0.142&0.627\cr
		PV-KITTI~\cite{angelina2018pointnetvlad}&0.882&0.791&0.734&0.953&0.767&0.129&0.709\cr
		Ours-RN&0.960&0.859&0.897&0.944&\textbf{0.984}&0.783&0.904\cr
		Ours-SK&\textbf{0.969}&\textbf{0.891}&0.905&0.971&0.967&\textbf{0.900}&\textbf{0.934}\cr
		\bottomrule
	\end{tabular}
	\vspace{5pt}
	\caption{$F_1$ max scores on KITTI dataset.}
	\label{table:F1_ori}
\end{table}

\subsection{Place Recognition Performance}

\begin{figure*}[h]
	\centerline{\includegraphics[width=4.8in]{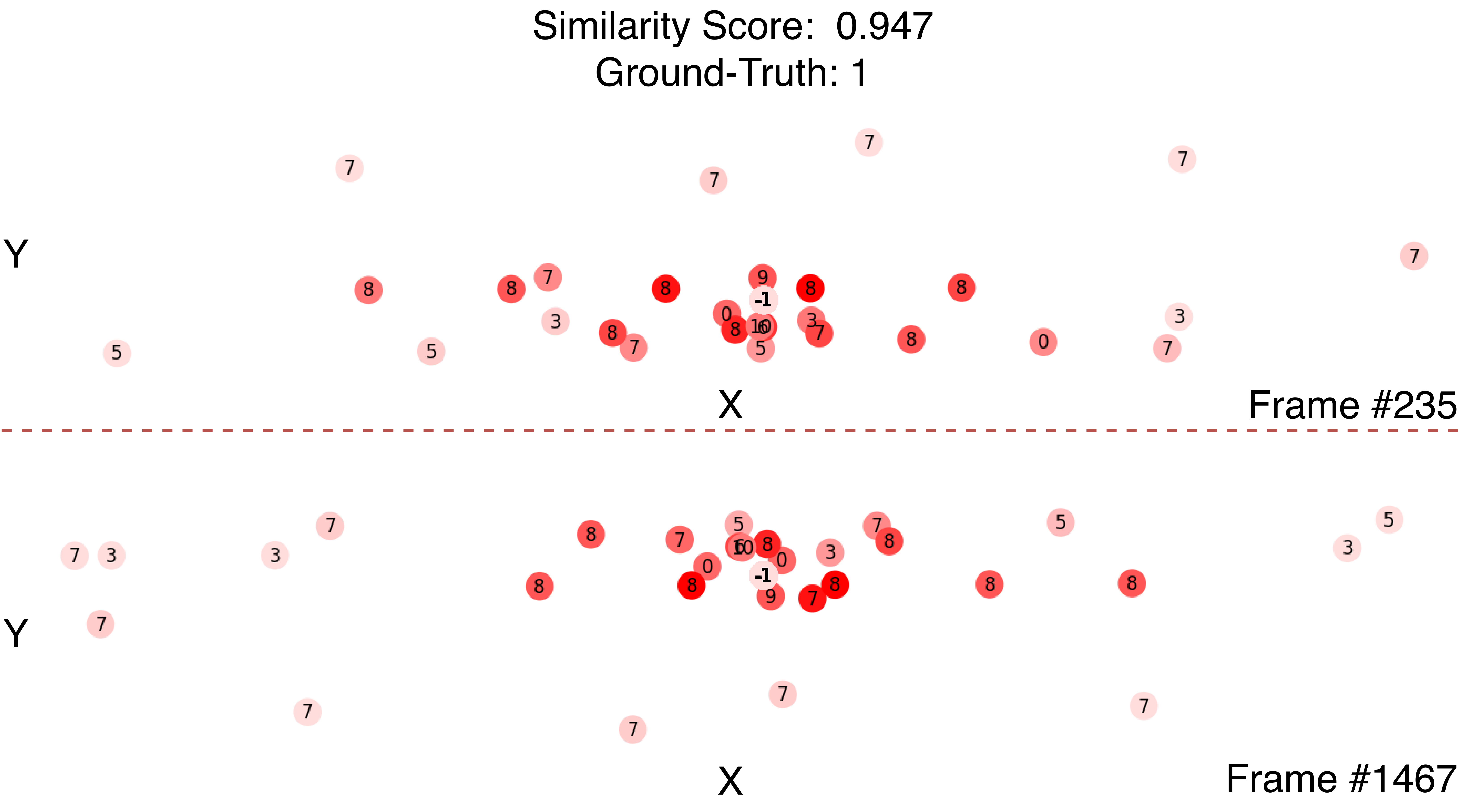}}
	\caption{Graph visualization with attention weights from KITTI08. Note that it's the same place with  opposite views. The number denotes the category of each node, and the color intensity represents the attention weight. The deeper the color, the higher the weight. Fake nodes (zero-padding nodes) are -1 in figure and their one-hot encoding is zero vector.} 
	\label{fig:att}
\end{figure*}

\begin{table*}[htb]
	\small
	\centering
	\setlength{\tabcolsep}{1.1mm}
	\begin{tabular}{ccccccccccccccccc}
		\toprule
		\multirow{2}{*}{Methods}&
		\multicolumn{8}{c}{Occlusion} & \multicolumn{8}{c}{Rotation} \cr
		\cmidrule(lr){2-9} \cmidrule(l){10-17}
		& 00 & 02 & 05 & 06 & 07 & 08 & Mean & Cmp & 00 & 02 & 05 & 06 & 07 & 08 & Mean & Cmp \cr
		\midrule
		M2DP\cite{he2016m2dp} &0.251&0.128&0.324&0.669&0.549&0.102&0.337 &-0.382 &0.425&0.344&0.415&0.668&0.590&0.348&0.465&-0.245\cr
		SC~\cite{Kim2018ScanCE}&0.916&\textbf{0.847}&\textbf{0.925}&\textbf{0.996}&0.850&0.721&0.876&-0.038
		&0.937&0.859&\textbf{0.954}&\textbf{0.998}&0.936&0.813&0.916&\textbf{+0.002}\cr
		PV-PRE~\cite{angelina2018pointnetvlad}&0.664&0.610&0.661&0.813&0.439&0.169&0.560&-0.067
		&0.332&0.133&0.348&0.668&0.647&0.202&0.388&-0.239\cr
		PV-KITTI~\cite{angelina2018pointnetvlad}&0.777&0.696&0.632&0.900&0.579&0.112&0.619&-0.090
		&0.253&0.132&0.713&0.670&0.435&0.156&0.393&-0.316\cr
		Ours-RN&0.935&0.817&0.862&0.932&0.928&0.754&0.871&\textbf{-0.033}
		&0.959&0.858&0.894&0.939&\textbf{0.977}&0.779&0.901&-0.003\cr
		Ours-SK&\textbf{0.941}&0.841&0.864&0.954&\textbf{0.935}&\textbf{0.844}&\textbf{0.897}&-0.037
		&\textbf{0.968}&\textbf{0.892}&0.902&0.966&0.965&\textbf{0.903}&\textbf{0.933}&-0.001\cr
		\bottomrule
	\end{tabular}
	\vspace{5pt}
	\caption{$F_1$ max scores on KITTI dataset when the point clouds are randomly occluded with $30^{\circ}$ and rotated around z-axis. Cmp is the comparison with the standard results shown in Table.~\ref{table:F1_ori}}
	\label{table:F1_robustness}
\end{table*}

To evaluate our semantic graph representation and graph similarity network, we use both the RangeNet++ predictions from model darknet53\footnote{\url{https://github.com/PRBonn/lidar-bonnetal/tree/master/train/tasks/semantic}} (Ours-RN) and SemanticKITTI label (Ours-SK) as front-end, comparing with M2DP\footnote{\url{https://github.com/LiHeUA/M2DP}}, Scan Context\footnote{\url{https://github.com/irapkaist/scancontext}} (SC) and PointNetVLAD. Specifically, for PointNetVLAD, we use both its pretrained (refined version) model\footnote{\url{https://drive.google.com/file/d/1wYsJmfd2yfbK9DHjFHwEeU1a_x35od61/view}} (PV-PRE) and retrained model on KITTI (PV-KITTI) taking advantage of 1-fold strategy.

\textbf{Quantitative Results:} We analyze the performance using the precision-recall curve in Fig.~\ref{Fig:PRcurve}, and we calculate the maximum value of $F_1$ score to evaluate different precision-recall curves shown in Table.~\ref{table:F1_ori}. $F_1$ score is defined as
\begin{equation}\label{eq:f1max}
F_1 = 2 \times \frac{P \times R}{P+R},
\end{equation}
where $P$ denotes precision and $R$ denotes recall. As shown in Fig.~\ref{Fig:PRcurve} and Table.~\ref{table:F1_ori}, our mean $F_1$ max score outperforms other existing methods and our overall performance is competitive. Especially for the challenging sequence 08 with reverse loops, M2DP, PV-PRE, and PV-KITTI have severe degradation. Such methods based on global descriptors cannot handle the viewpoint variations, while our approach performs consistently. Thus when the viewpoint changes, we can still report a confident result, which is further proved in Section~\ref{sec:robustness test}. Notably, the IoU (intersection-over-union) of RangeNet++ is only 52\%, which is not high and will introduce noises like wrong labels and centroid offset. Though Ours-RN is indeed lower than Ours-SK, it performs satisfactorily, which indicates that better semantic prediction will bring improvements. The evaluation results demonstrate that our method is effective in large-scale place recognition.

\textbf{Qualitative Results:} Visualizations of the node attentions in graph embedding is shown in Fig.~\ref{fig:att}. We observe that nodes closer to the center receive higher attention while fake nodes (the zero-padding nodes are -1 in figure and their one-hot encoding is zero vector) receive lower attention. These results are of intuitive significance and further prove the effectiveness of the proposed method.

\begin{figure}[t]
	\centerline{\includegraphics[width=3.0in]{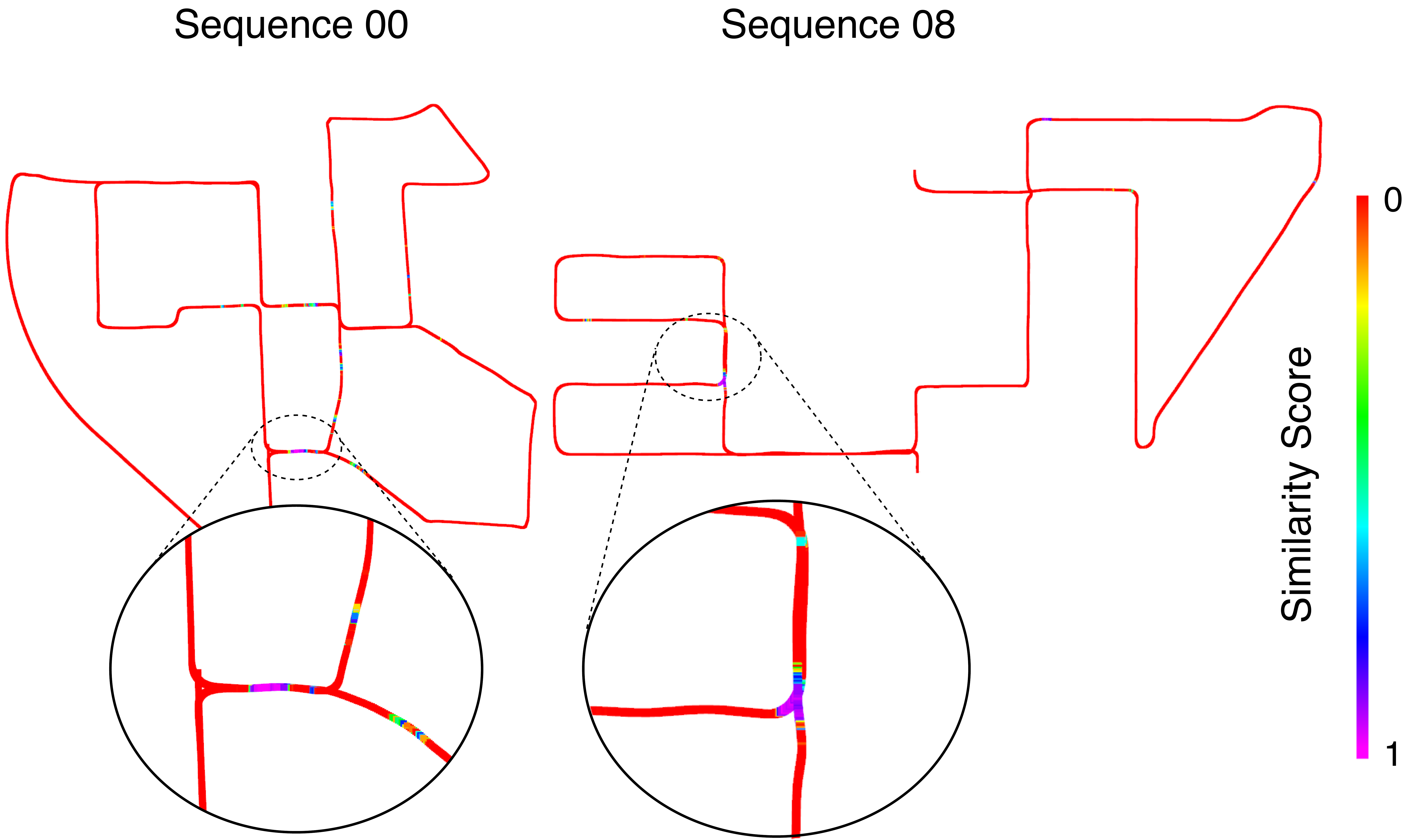}}
	\caption{Similarity visualization. We randomly choose a single scene in each sequence and zoom in similarity scores around this scene. (Best viewed with zoom-in.)} 
	\label{fig:detail}
\end{figure}

Furthermore, we show a piece of qualitative visualization results in Fig.~\ref{fig:detail}. We randomly choose a specific single scene and measure similarities between it with other scenes in each sequence. Then, we color the trajectories according to scores. In fact, only the scenes near this chosen scene
within a distance threshold (e.g. 3 m) should be similar (color closer to purple). We find our predictions are distinct and accurate.

\subsection{Robustness Test}
\label{sec:robustness test}
\smallskip
\textbf{Occlusion:} In a real scenario, dynamic objects (person, vehicle, etc.) inevitably occur during long-term localization, which brings occlusion in LiDAR point clouds. For validation, we randomly select a certain angle in the azimuth and remove points within this area. In practice, we report results with a removal range of $30^{\circ}$ in Table.~\ref{table:F1_robustness}. All methods have some performance degradation due to information loss, while our methods are the least affected. Because M2DP depends on the point projection, it is sensitive to the change of point distribution caused by the occlusion. PointNetVLAD randomly samples 4096 points per submap as its input, resulting in vulnerability to the absence of the raw point clouds. Scan Context divides the scene into several bins and the miss of a few bins has less impact. The occlusion in our semantic graph representation is equivalent to the disappearance of some nodes, which is only a small part of our representation, and the graph similarity network is robust against the missing to some extent.

\smallskip
\textbf{Viewpoint Changes:} In practice, the viewpoint often changes when arriving at the same place. Thus it‘s crucial to handle the viewpoint changes. We randomly rotate the point cloud’s heading and the results are shown in Table.~\ref{table:F1_robustness}. Due to the lack of rotation invariance, the effectiveness of most methods drops dramatically. Although PointNetVLAD has the T-Net modules aiming to regularize the input, it is still not enough to achieve the rotation invariance. Scan Context calculates distances with all possible column-shifted scan contexts and finds the minimum distance, which introduces repeat computation. Compared with local features and distributed features, the semantic information and topological relations among nodes are rotation invariant and our semantic graph representation captures both of them. The spatial and feature relations among nodes are encoded in the node embedding part of our graph similarity network. Besides, compared with the raw point clouds, the number of nodes in the graph is relatively small, which reduces solution space and enables the network easy to converge.

\begin{figure}[tb]
	\centering
	\begin{subfigure}{0.5\textwidth}
		\centering
		\includegraphics[width=0.8\textwidth]{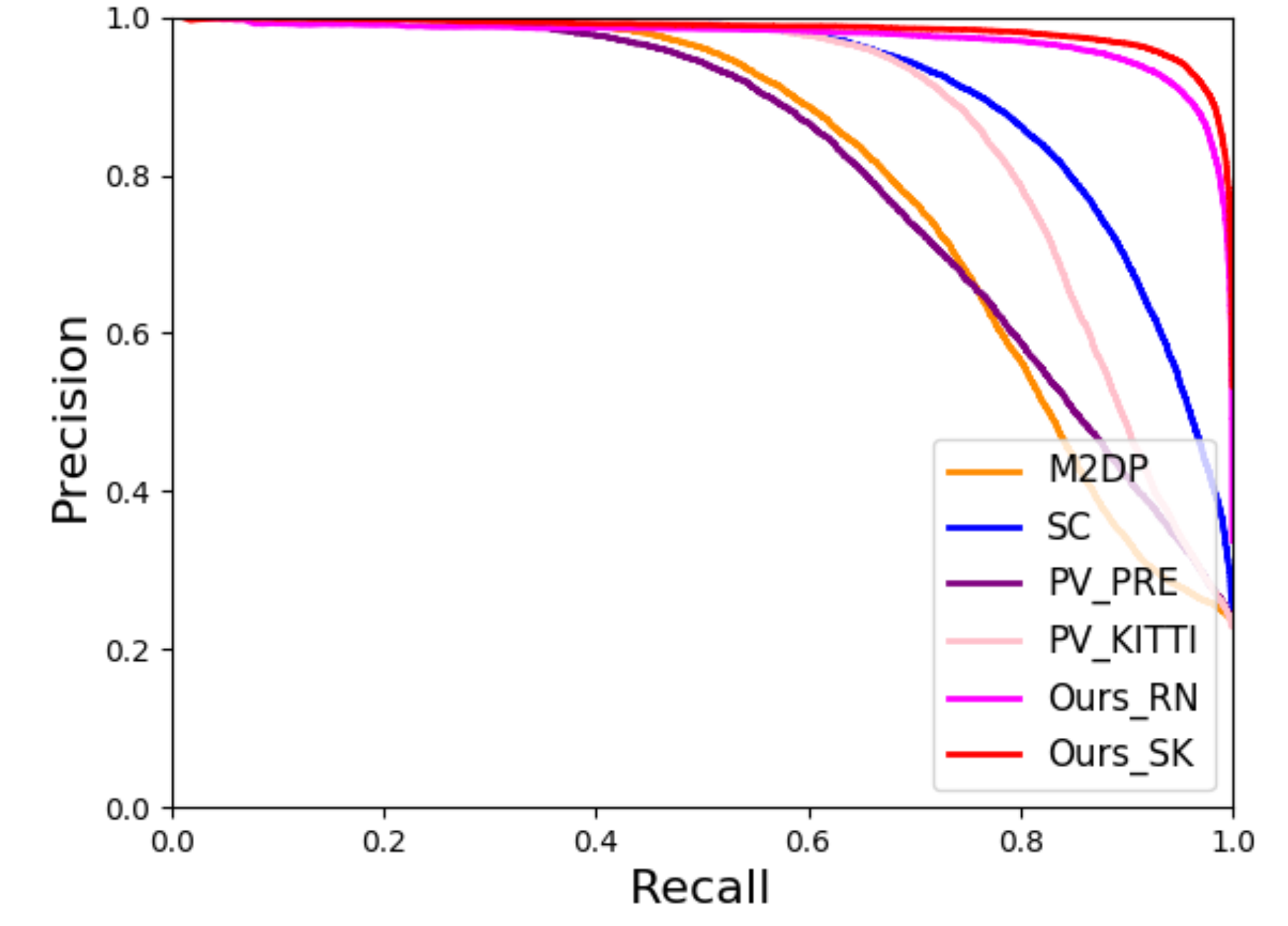}
		\caption{5 m}
		\label{fig:KITTI00_5m}
	\end{subfigure}

	\begin{subfigure}{0.5\textwidth}
		\centering
		\includegraphics[width=0.8\textwidth]{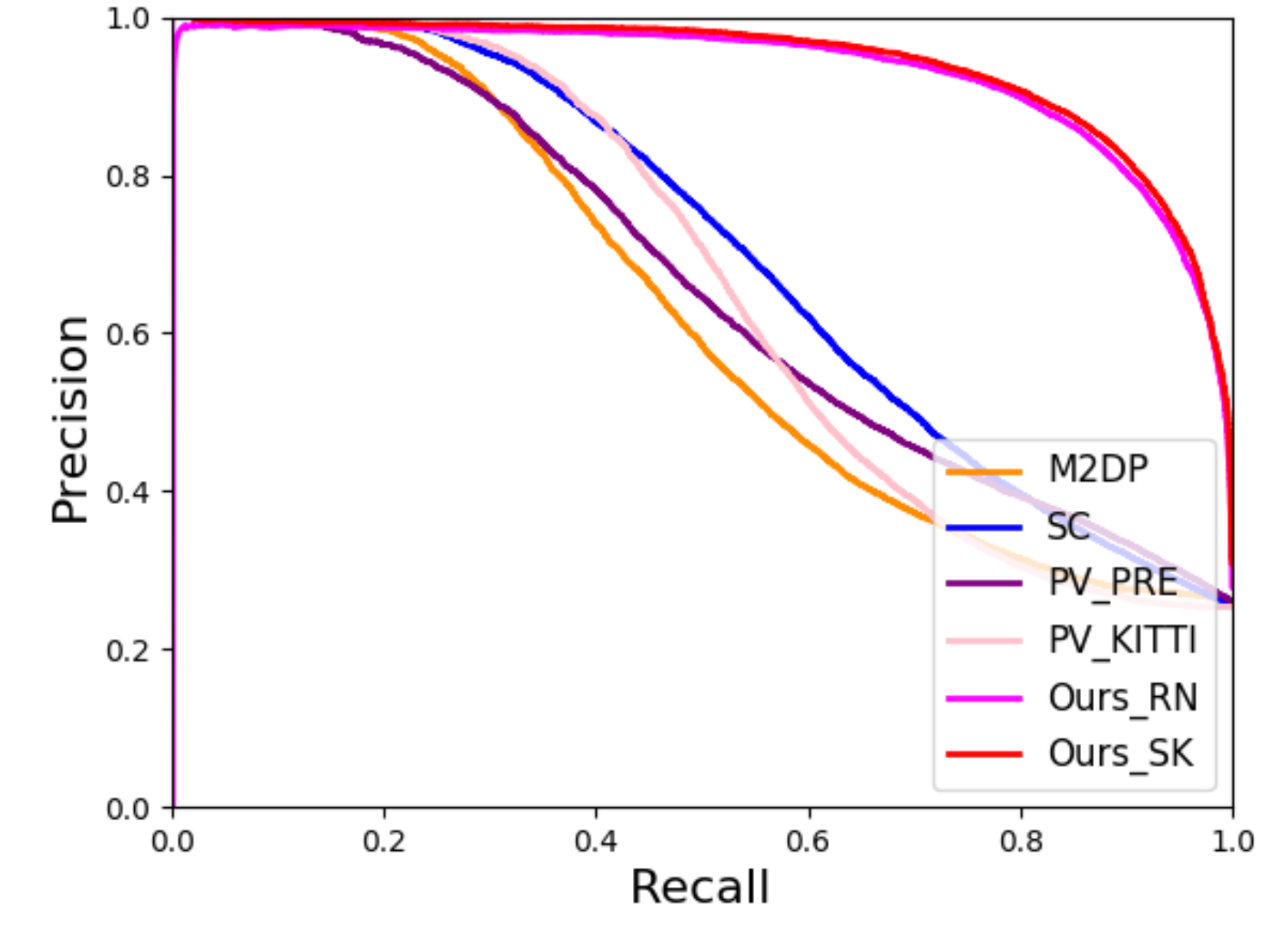}
		\caption{10 m}
		\label{fig:KITTI00_10m}
	\end{subfigure}

	\centering
	\caption{Precision-Recall curves on KITTI00 with different distance thresholds.}
	\label{Fig:threshold}
\end{figure}

\smallskip
\textbf{Distance Thresholds:} For specific task and application scenario, different distance thresholds of positive pairs for loop closure is needed. To evaluate the adaptability of our approach, we adopt thresholds of 3 m, 5 m, and 10 m to train the corresponding models. The Precision-Recall curves on KITTI00 with 3 m, 5 m and 10 m are shown in Fig.~\ref{fig:KITTI00_3m}, Fig.~\ref{fig:KITTI00_5m} and Fig.~\ref{fig:KITTI00_10m} respectively, which indicates that our approach can maintain satisfactory performance.

\subsection{Efficiency}
For each frame, the descriptor size of M2DP, PointNetVLAD, Scan Context and ours is 192, 256, 20$\times$60 and 100$\times$4 (N$\times$$f_i$), respectively. Our graph similarity network is lightweight, fast, and capable of obtaining similarity scores for $N$ places via a single pass through the network instead of comparing them one by one. With the batch size of 128, the inference takes about 9 ms and occupies 2820 MB GPU memory on an NVIDIA GeForce RTX 2080 Ti, making it applicable in real-time robotics systems.

\section{Conclusion}
 
In this paper, we propose a novel semantic graph based approach for large-scale place recognition in 3D point clouds, providing a promising direction for future research for a more complete exploitation of semantic information for place recognition. Compared to the existing methods focus on extracting local, global, and statistical features of raw point clouds, acting at the semantic level offers several advantages in environmental changes and is more closely to the way humans perceive scenes. Exhaustive evaluations demonstrate the feasibility and robustness of our approach, especially for reverse loops. 

In future work, we will investigate in unsupervised semantic feature learning of point clouds and its availability on place recognition. 

\addtolength{\topmargin}{0.249cm}
\bibliographystyle{ieeetr}
\bibliography{root}

\begin{thebibliography}{10}

\bibitem{cummins2008fab}
M.~Cummins and P.~Newman, ``Fab-map: Probabilistic localization and mapping in
  the space of appearance,'' {\em The International Journal of Robotics
  Research}, vol.~27, no.~6, pp.~647--665, 2008.

\bibitem{milford2012seqslam}
M.~J. Milford and G.~F. Wyeth, ``Seqslam: Visual route-based navigation for
  sunny summer days and stormy winter nights,'' in {\em 2012 IEEE International
  Conference on Robotics and Automation}, pp.~1643--1649, IEEE, 2012.

\bibitem{sattler2016large}
T.~Sattler, M.~Havlena, K.~Schindler, and M.~Pollefeys, ``Large-scale location
  recognition and the geometric burstiness problem,'' in {\em Proceedings of
  the IEEE Conference on Computer Vision and Pattern Recognition},
  pp.~1582--1590, 2016.

\bibitem{lazebnik2006beyond}
S.~Lazebnik, C.~Schmid, and J.~Ponce, ``Beyond bags of features: Spatial
  pyramid matching for recognizing natural scene categories,'' in {\em 2006
  IEEE Computer Society Conference on Computer Vision and Pattern Recognition
  (CVPR'06)}, vol.~2, pp.~2169--2178, IEEE, 2006.

\bibitem{jegou2010aggregating}
H.~J{\'e}gou, M.~Douze, C.~Schmid, and P.~P{\'e}rez, ``Aggregating local
  descriptors into a compact image representation,'' in {\em CVPR 2010-23rd
  IEEE Conference on Computer Vision \& Pattern Recognition}, pp.~3304--3311,
  IEEE Computer Society, 2010.

\bibitem{dixit2015scene}
M.~Dixit, S.~Chen, D.~Gao, N.~Rasiwasia, and N.~Vasconcelos, ``Scene
  classification with semantic fisher vectors,'' in {\em Proceedings of the
  IEEE conference on computer vision and pattern recognition}, pp.~2974--2983,
  2015.

\bibitem{dixit2016object}
M.~D. Dixit and N.~Vasconcelos, ``Object based scene representations using
  fisher scores of local subspace projections,'' in {\em Advances in Neural
  Information Processing Systems}, pp.~2811--2819, 2016.

\bibitem{lowry2015visual}
S.~Lowry, N.~S{\"u}nderhauf, P.~Newman, J.~J. Leonard, D.~Cox, P.~Corke, and
  M.~J. Milford, ``Visual place recognition: A survey,'' {\em IEEE Transactions
  on Robotics}, vol.~32, no.~1, pp.~1--19, 2015.

\bibitem{he2016m2dp}
L.~He, X.~Wang, and H.~Zhang, ``M2dp: A novel 3d point cloud descriptor and its
  application in loop closure detection,'' in {\em 2016 IEEE/RSJ International
  Conference on Intelligent Robots and Systems (IROS)}, pp.~231--237, IEEE,
  2016.

\bibitem{angelina2018pointnetvlad}
M.~Angelina~Uy and G.~Hee~Lee, ``Pointnetvlad: Deep point cloud based retrieval
  for large-scale place recognition,'' in {\em Proceedings of the IEEE
  Conference on Computer Vision and Pattern Recognition}, pp.~4470--4479, 2018.

\bibitem{liu2019lpd}
Z.~Liu, S.~Zhou, C.~Suo, P.~Yin, W.~Chen, H.~Wang, H.~Li, and Y.-H. Liu,
  ``Lpd-net: 3d point cloud learning for large-scale place recognition and
  environment analysis,'' in {\em Proceedings of the IEEE International
  Conference on Computer Vision}, pp.~2831--2840, 2019.

\bibitem{liu2019seqlpd}
Z.~Liu, C.~Suo, S.~Zhou, F.~Xu, H.~Wei, W.~Chen, H.~Wang, X.~Liang, and Y.-H.
  Liu, ``Seqlpd: Sequence matching enhanced loop-closure detection based on
  large-scale point cloud description for self-driving vehicles,'' in {\em 2019
  IEEE/RSJ International Conference on Intelligent Robots and Systems (IROS)},
  pp.~1218--1223, IEEE, 2019.

\bibitem{dube2017segmatch}
R.~Dub{\'e}, D.~Dugas, E.~Stumm, J.~Nieto, R.~Siegwart, and C.~Cadena,
  ``Segmatch: Segment based place recognition in 3d point clouds,'' in {\em
  2017 IEEE International Conference on Robotics and Automation (ICRA)},
  pp.~5266--5272, IEEE, 2017.

\bibitem{segmap2018}
R.~Dub{\'e}, A.~Cramariuc, D.~Dugas, J.~Nieto, R.~Siegwart, and C.~Cadena,
  ``{SegMap}: 3d segment mapping using data-driven descriptors,'' in {\em
  Robotics: Science and Systems (RSS)}, 2018.

\bibitem{dube2019segmap}
R.~Dub{\'e}, A.~Cramariuc, D.~Dugas, H.~Sommer, M.~Dymczyk, J.~Nieto,
  R.~Siegwart, and C.~Cadena, ``Segmap: Segment-based mapping and localization
  using data-driven descriptors,'' {\em The International Journal of Robotics
  Research}, p.~0278364919863090, 2019.

\bibitem{geiger2013vision}
A.~Geiger, P.~Lenz, C.~Stiller, and R.~Urtasun, ``Vision meets robotics: The
  kitti dataset,'' {\em The International Journal of Robotics Research},
  vol.~32, no.~11, pp.~1231--1237, 2013.

\bibitem{johnson1999using}
A.~E. Johnson and M.~Hebert, ``Using spin images for efficient object
  recognition in cluttered 3d scenes,'' {\em IEEE Transactions on pattern
  analysis and machine intelligence}, vol.~21, no.~5, pp.~433--449, 1999.

\bibitem{bosse2013place}
M.~Bosse and R.~Zlot, ``Place recognition using keypoint voting in large 3d
  lidar datasets,'' in {\em 2013 IEEE International Conference on Robotics and
  Automation}, pp.~2677--2684, IEEE, 2013.

\bibitem{wohlkinger2011ensemble}
W.~Wohlkinger and M.~Vincze, ``Ensemble of shape functions for 3d object
  classification,'' in {\em 2011 IEEE international conference on robotics and
  biomimetics}, pp.~2987--2992, IEEE, 2011.

\bibitem{wang2019lidar}
Y.~Wang, Z.~Sun, J.~Yang, and H.~Kong, ``Lidar iris for loop-closure
  detection,'' {\em arXiv preprint arXiv:1912.03825}, 2019.

\bibitem{qi2017pointnet}
C.~R. Qi, H.~Su, K.~Mo, and L.~J. Guibas, ``Pointnet: Deep learning on point
  sets for 3d classification and segmentation,'' in {\em Proceedings of the
  IEEE Conference on Computer Vision and Pattern Recognition}, pp.~652--660,
  2017.

\bibitem{arandjelovic2016netvlad}
R.~Arandjelovic, P.~Gronat, A.~Torii, T.~Pajdla, and J.~Sivic, ``Netvlad: Cnn
  architecture for weakly supervised place recognition,'' in {\em Proceedings
  of the IEEE conference on computer vision and pattern recognition},
  pp.~5297--5307, 2016.

\bibitem{Kim2018ScanCE}
G.~Kim and A.~Kim, ``Scan context: Egocentric spatial descriptor for place
  recognition within 3d point cloud map,'' {\em 2018 IEEE/RSJ International
  Conference on Intelligent Robots and Systems (IROS)}, pp.~4802--4809, 2018.

\bibitem{Wu2017SqueezeSegCN}
B.~Wu, A.~Wan, X.~Yue, and K.~Keutzer, ``Squeezeseg: Convolutional neural nets
  with recurrent crf for real-time road-object segmentation from 3d lidar point
  cloud,'' {\em 2018 IEEE International Conference on Robotics and Automation
  (ICRA)}, pp.~1887--1893, 2017.

\bibitem{wu2019squeezesegv2}
B.~Wu, X.~Zhou, S.~Zhao, X.~Yue, and K.~Keutzer, ``Squeezesegv2: Improved model
  structure and unsupervised domain adaptation for road-object segmentation
  from a lidar point cloud,'' in {\em 2019 International Conference on Robotics
  and Automation (ICRA)}, pp.~4376--4382, IEEE, 2019.

\bibitem{Kong2019PASS3DPA}
X.~Kong, G.~Zhai, B.~Zhong, and Y.~Liu, ``Pass3d: Precise and accelerated
  semantic segmentation for 3d point cloud,'' {\em 2019 IEEE/RSJ International
  Conference on Intelligent Robots and Systems (IROS)}, pp.~3467--3473, 2019.

\bibitem{milioto2019rangenet++}
A.~Milioto, I.~Vizzo, J.~Behley, and C.~Stachniss, ``Rangenet++: Fast and
  accurate lidar semantic segmentation,'' in {\em Proc. of the IEEE/RSJ Intl.
  Conf. on Intelligent Robots and Systems (IROS)}, 2019.

\bibitem{behley2019semantickitti}
J.~Behley, M.~Garbade, A.~Milioto, J.~Quenzel, S.~Behnke, C.~Stachniss, and
  J.~Gall, ``Semantickitti: A dataset for semantic scene understanding of lidar
  sequences,'' in {\em Proceedings of the IEEE International Conference on
  Computer Vision}, pp.~9297--9307, 2019.

\bibitem{yin2018locnet}
H.~Yin, L.~Tang, X.~Ding, Y.~Wang, and R.~Xiong, ``Locnet: Global localization
  in 3d point clouds for mobile vehicles,'' in {\em 2018 IEEE Intelligent
  Vehicles Symposium (IV)}, pp.~728--733, IEEE, 2018.

\bibitem{bunke1983distance}
H.~Bunke, ``What is the distance between graphs,'' {\em Bulletin of the EATCS},
  vol.~20, pp.~35--39, 1983.

\bibitem{bunke1998graph}
H.~Bunke and K.~Shearer, ``A graph distance metric based on the maximal common
  subgraph,'' {\em Pattern recognition letters}, vol.~19, no.~3-4,
  pp.~255--259, 1998.

\bibitem{zeng2009comparing}
Z.~Zeng, A.~K. Tung, J.~Wang, J.~Feng, and L.~Zhou, ``Comparing stars: On
  approximating graph edit distance,'' {\em Proceedings of the VLDB Endowment},
  vol.~2, no.~1, pp.~25--36, 2009.

\bibitem{blumenthal2018exact}
D.~B. Blumenthal and J.~Gamper, ``On the exact computation of the graph edit
  distance,'' {\em Pattern Recognition Letters}, 2018.

\bibitem{bai2019simgnn}
Y.~Bai, H.~Ding, S.~Bian, T.~Chen, Y.~Sun, and W.~Wang, ``Simgnn: A neural
  network approach to fast graph similarity computation,'' in {\em Proceedings
  of the Twelfth ACM International Conference on Web Search and Data Mining},
  pp.~384--392, ACM, 2019.

\bibitem{kipf2017semi}
T.~N. Kipf and M.~Welling, ``Semi-supervised classification with graph
  convolutional networks,'' in {\em International Conference on Learning
  Representations (ICLR)}, 2017.

\bibitem{wang2019dynamic}
Y.~Wang, Y.~Sun, Z.~Liu, S.~E. Sarma, M.~M. Bronstein, and J.~M. Solomon,
  ``Dynamic graph cnn for learning on point clouds,'' {\em ACM Transactions on
  Graphics (TOG)}, vol.~38, no.~5, p.~146, 2019.

\bibitem{husain2016combining}
F.~Husain, H.~Schulz, B.~Dellen, C.~Torras, and S.~Behnke, ``Combining semantic
  and geometric features for object class segmentation of indoor scenes,'' {\em
  IEEE Robotics and Automation Letters}, vol.~2, no.~1, pp.~49--55, 2016.

\bibitem{bao2019spatiotemporal}
J.~Bao, P.~Liu, and S.~V. Ukkusuri, ``A spatiotemporal deep learning approach
  for citywide short-term crash risk prediction with multi-source data,'' {\em
  Accident Analysis \& Prevention}, vol.~122, pp.~239--254, 2019.

\bibitem{geng2019spatiotemporal}
X.~Geng, Y.~Li, L.~Wang, L.~Zhang, Q.~Yang, J.~Ye, and Y.~Liu, ``Spatiotemporal
  multi-graph convolution network for ride-hailing demand forecasting,'' in
  {\em Proceedings of the AAAI Conference on Artificial Intelligence}, vol.~33,
  pp.~3656--3663, 2019.

\bibitem{socher2013reasoning}
R.~Socher, D.~Chen, C.~D. Manning, and A.~Ng, ``Reasoning with neural tensor
  networks for knowledge base completion,'' in {\em Advances in neural
  information processing systems}, pp.~926--934, 2013.

\bibitem{paszke2019pytorch}
A.~Paszke, S.~Gross, F.~Massa, A.~Lerer, J.~Bradbury, G.~Chanan, T.~Killeen,
  Z.~Lin, N.~Gimelshein, L.~Antiga, {\em et~al.}, ``Pytorch: An imperative
  style, high-performance deep learning library,'' in {\em Advances in Neural
  Information Processing Systems}, pp.~8024--8035, 2019.

\bibitem{kingma2014adam}
D.~P. Kingma and J.~Ba, ``Adam: A method for stochastic optimization,'' {\em
  arXiv preprint arXiv:1412.6980}, 2014.

\end{thebibliography}

\end{document}